\documentclass[sigconf,natbib=true,anonymous=false]{acmart}

\AtBeginDocument{%
  \providecommand\BibTeX{{%
    \normalfont B\kern-0.5em{\scshape i\kern-0.25em b}\kern-0.8em\TeX}}}

\setcopyright{acmcopyright}
\copyrightyear{2018}
\acmYear{2018}
\acmDOI{10.1145/1122445.1122456}

\acmConference[Woodstock '18]{Woodstock '18: ACM Symposium on Neural
  Gaze Detection}{June 03--05, 2018}{Woodstock, NY}
\acmBooktitle{Woodstock '18: ACM Symposium on Neural Gaze Detection,
  June 03--05, 2018, Woodstock, NY}
\acmPrice{15.00}
\acmISBN{978-1-4503-XXXX-X/18/06}

\usepackage{graphicx}
\usepackage{amsmath}

\usepackage{amssymb}
\usepackage{booktabs}

\usepackage{amsfonts}
\usepackage{bbm}
\usepackage{multirow}
\usepackage[ruled,linesnumbered]{algorithm2e}
\usepackage[noend]{algpseudocode}
\begin{document}

\title{CLS: Cross Labeling Supervision for Semi-Supervised Learning}

% \author {
%     % Authors
%     Yao Yao,\textsuperscript{\rm 1, 2}
%     % Wanpeng Zhang, \textsuperscript{\rm 3}
%     Junyi Shen, \textsuperscript{\rm 2}
%     Jin Xu, \textsuperscript{\rm 2}\thanks{corresponding authors.}
%     Bin Zhong, \textsuperscript{\rm 2}
%     Li Xiao, \textsuperscript{\rm 1*}
% }
% \affiliations {
%     % Affiliations
%     \textsuperscript{\rm 1} TBSI, Tsinghua Shenzhen International Graduate School, Tsinghua University, China\\
%     \textsuperscript{\rm 2} Data Quality Team, WeChat, Tencent Inc., China\\
%     % \textsuperscript{\rm 3} Informatics Division, Tsinghua Shenzhen International Graduate School, Tsinghua University, China\\
%     \{y-yao19@mails, xiaoli@sz\}.tsinghua.edu.cn, \{vichyshen, jinxxu, harryzhong\}@tencent.com
%     % firstAuthor@affiliation1.com, secondAuthor@affilation2.com, thirdAuthor@affiliation1.com
% }
\author{Yao Yao}
\affiliation{%
  \institution{TBSI, Tsinghua Shenzhen International Graduate School, Tsinghua University}
  \city{Shenzhen}
  \country{China}
}
\email{y-yao19@mails.tsinghua.edu.cn}

\author{Junyi Shen}
\affiliation{%
  \institution{Data Quality Team, WeChat, \\ Tencent Inc.}
  \city{Shenzhen}
  \country{China}}
\email{vichyshen@tencent.com}

\author{Jin Xu}
\affiliation{%
  \institution{Data Quality Team, WeChat, \\ Tencent Inc.}
  \city{Shenzhen}
  \country{China}}
\email{jinxxu@tencent.com}

\author{Bin Zhong}
\affiliation{%
  \institution{Data Quality Team, WeChat, \\ Tencent Inc.}
  \city{Shenzhen}
  \country{China}}
\email{harryzhong@tencent.com}

\author{Li Xiao}
\affiliation{%
  \institution{TBSI, Tsinghua Shenzhen International Graduate School, Tsinghua University}
  \city{Shenzhen}
  \country{China}
}
\email{xiaoli@sz.tsinghua.edu.cn}

% \affiliation{%

% \affiliation{%

% \affiliation{%

% \affiliation{%

% \affiliation{%

% \affiliation{%

\renewcommand{\shortauthors}{Yao and Junyi, et al.}

%%%%%%%%% ABSTRACT
\begin{abstract}
It is well known that the success of deep neural networks is greatly attributed to large-scale labeled datasets. However, it can be extremely time-consuming and laborious to collect sufficient high-quality labeled data in most practical applications. Semi-supervised learning (SSL) provides an effective solution to reduce the cost of labeling by simultaneously leveraging both labeled and unlabeled data. In this work, we present Cross Labeling Supervision (CLS), a framework that generalizes the typical pseudo-labeling process. Based on FixMatch~\cite{sohn2020fixmatch}, where a pseudo label is generated from a weakly-augmented sample to teach the prediction on a strong augmentation of the same input sample, CLS allows the creation of both pseudo and complementary labels to support both positive and negative learning. To mitigate the confirmation bias of self-labeling and boost the tolerance to false labels, two different initialized networks with the same structure are trained simultaneously. Each network utilizes high-confidence labels from the other network as additional supervision signals. During the label generation phase, adaptive sample weights are assigned to artificial labels according to their prediction confidence. The sample weight plays two roles: quantify the generated labels' quality and reduce the disruption of inaccurate labels on network training. Experimental results on the semi-supervised classification task show that our framework outperforms existing approaches by large margins on the CIFAR-10 and CIFAR-100 datasets.
\end{abstract}

\begin{CCSXML}
<ccs2012>
   <concept>
       <concept_id>10010147.10010178.10010224.10010245</concept_id>
       <concept_desc>Computing methodologies~Computer vision problems</concept_desc>
       <concept_significance>500</concept_significance>
       </concept>
   <concept>
       <concept_id>10010147.10010257.10010282.10011305</concept_id>
       <concept_desc>Computing methodologies~Semi-supervised learning settings</concept_desc>
       <concept_significance>500</concept_significance>
       </concept>
 </ccs2012>
\end{CCSXML}

\ccsdesc[500]{Computing methodologies~Computer vision problems}
\ccsdesc[500]{Computing methodologies~Semi-supervised learning settings}

\keywords{semi-supervised learning, pseudo label, image classification}


\maketitle
      % To produce the REVIEW version
%\usepackage{cvpr}              % To produce the CAMERA-READY version
%\usepackage[pagenumbers]{cvpr} % To force page numbers, e.g. for an arXiv version

% Include other packages here, before hyperref.


%%%%%%%%% BODY TEXT
\section{Introduction}
\label{sec:intro}
% 基准方法：Fixmatch
% intro：传统的self-training 存在一个致命问题，在训练过程中出现的预测错误会被不断累积和放大，最终导致这一错误认知被不断强化和固定下来，这一现象叫作确认偏差（confirmation bias）。为了解决这一问题，我们引入独立解耦合的同行网络（peer network）来进行伪标签的生成。为了保证模型的互补性，以及避免模型坍塌（即两个模型同质化），我们对两个模型使用不同的初始化， 
% The recent extraordinary success of deep learning methods in computer vision tasks cannot be separated from the collection of large-scale labeled datasets, such as ImageNet~\cite{russakovsky2015imagenet} and WebVision~\cite{li2017webvision}. However, it can be fairly expensive and time-consuming to get high-quality labels through manual annotation. To alleviate the dependence on huge labeled datasets, semi-supervised learning (SSL) has gained more and more attention and become an active research area due to its desired ability to effectively exploit unlabeled data.
% Since unlabeled data can often be obtained at low cost, SSL has demonstrated superior performance on various tasks such as semantic segmentation~\cite{chen2021semi}, image classification~\cite{sohn2020fixmatch}, and object detection~\cite{yang2021interactive}.
The recent extraordinary success of deep learning methods in computer vision tasks cannot be separated from the collection of large-scale labeled datasets, such as ImageNet~\cite{russakovsky2015imagenet} and WebVision~\cite{li2017webvision}. However, it can be pretty expensive and time-consuming to get high-quality labels through manual annotation. To alleviate the dependence on massive labeled datasets, semi-supervised learning (SSL) has gained more and more attention and become an active research area due to its desired ability to exploit unlabeled data effectively.
Since unlabeled data can often be obtained at low cost, SSL has demonstrated superior performance on various tasks such as semantic segmentation~\cite{chen2021semi}, image classification~\cite{sohn2020fixmatch}, and object detection~\cite{yang2021interactive}.

\begin{figure}[t]
\centering
\includegraphics[width=0.47\textwidth]{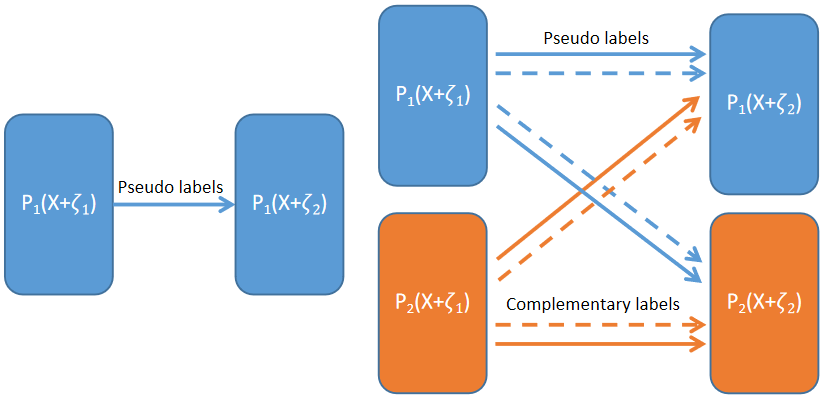}
\caption{
The difference between FixMatch~\cite{sohn2020fixmatch} and CLS. 
\textbf{Left:} FixMatch, where a prediction on the weakly-augmented sample ($X+\zeta_1$) generates a pseudo label for a strong augmentation of the same sample ($X+\zeta_2$) to learn from. 
\textbf{Right:} CLS, where two parallel networks generate both pseudo labels and complementary labels for themselves and each other. 
In some sense, supervision from the other network serves as a regularizer to mitigate the confirmation bias, while complementary labels support negative learning to boost performance further.
% Supervision from the other network in some sense serves as a regularizer to mitigate the confirmation bias, while complementary labels support negative learning to further boost performance. 
}
\label{fig:diff}
\end{figure}

Researches on SSL usually start from some intuitive assumptions like \textit{smoothness}, \textit{low-density}, etc. For example, based on the \textit{smoothness assumption} --- ``If two data points in a high-density region are close, then so should be the corresponding outputs''~\cite{chapelle2009semi},  consistency-based methods impose different consistency constraints over augmented inputs or perturbed networks~\cite{srivastava2014dropout} to enforce that the model prediction is invariant against data augmentations and proximity in the latent space.  In this type of approaches~\cite{sajjadi2016regularization, miyato2018virtual, xie2020unsupervised}, the Teacher-Student structure is commonly used explicitly or implicitly to force the student to produce an output consistent with the teacher for the perturbed inputs. 
Similarly, low-entropy (i.e., high-confidence) regularization is employed to  meet the \textit{low-density assumption}. 
To encourage low-density separation between classes~\cite{grandvalet2004semi},
% and to push the decision boundary away from the labeled data points~\cite{grandvalet2004semi}, 
Pseudo-Label~\cite{lee2013pseudo}, a self-training approach~\cite{mcclosky2006effective,mukherjee2020uncertainty,xie2020self}, 
generates pseudo labels for unlabeled images based on the model's class predictions and then selects high-confidence pairs (unlabeled images and their pseudo labels) to expand the training set.
Self-training methods can be interpreted as a particular case of the Teacher-Student paradigm, where the teacher is a particular function of the student according to some predefined rules. Such rules include directly copying the student's parameters~\cite{rasmus2015semi}, adopting an exponential moving average of the previous iterations~\cite{laine2016temporal,tarvainen2017mean}, or designing a specific loss to optimize the teacher's parameters~\cite{pham2021meta}.
Furthermore, building on the respective advances in pseudo-labeling and consistency-based methods, current state-of-the-art methods tend to be a combination of these two types of methods. FixMatch~\cite{sohn2020fixmatch} uses the pseudo label of a weakly-augmented image to supervise the prediction of the same image under strong augmentation. MixMatch~\cite{berthelot2019mixmatch} averages the estimations on multiple augmentations and produces the training target of the consistency regularization by applying a temperature sharpening function over the estimations.

Despite the strong performance of self-training methods and consistency-based methods, they all suffer from the problem of confirmation bias~\cite{arazo2020pseudo}, that is, if the teacher/pseudo labels are inaccurate, training the student/model itself under the misleading guidance may lead to significant performance  degradation~\cite{tarvainen2017mean}. To alleviate this problem, we propose a framework, namely cross labeling supervision (CLS), which contains three modifications to FixMatch~\cite{sohn2020fixmatch}. 
(1) The first is the generation of complementary labels to support negative learning~\cite{kim2019nlnl}, which reduces the risk of providing wrong information since the chance of selecting the ground truth label as a complementary label is relatively low. Empirically, the additional generated complementary labels can help to calibrate incorrect predictions compared to only using pseudo labels as supervised signals. For ease of expression, pseudo labels and complementary labels are collectively referred to as artificial labels in this paper.
(2) The second is a sample re-weighting mechanism to down-weight low-confidence artificial labels.
Unlike FixMatch~\cite{sohn2020fixmatch}, which sets a confidence threshold to completely filter out low-confidence pseudo labels,
the advantage of using soft re-weighting is that the network can still learn from low-confidence labels for better generalization.
(3) Inspired by co-training~\cite{blum1998combining}, we propose to exploit the disagreement of two independent models to achieve a complementary effect and avoid memorizing the inaccurate self-labeling samples.
Specifically, two identical networks with different initializations are trained simultaneously, and they mitigate the confirmation bias of self-labeling by exchanging high-confidence artificial labels. 
Thanks to the development of parallel technology, training two networks simultaneously adds little computational overhead. 
Figure \ref{fig:diff} demonstrates the difference between FixMatch~\cite{sohn2020fixmatch} and CLS.
% Inspired by the effectiveness of negative learning for  noisy labels~\cite{kim2019nlnl}, the two models not only provide each other with high confident pseudo labels, but also generate negative complementary labels to indicate which categories the input images do not belong to. 

%  The benefits from the our Cross Labeling structure lie in two-fold. On the one hand, 
% % 一方面，在结合打标签和一致性正则化的优势的同时，通过co-training的方式减少确认偏差。另一方面，我们的方法易于实现和拓展，得益于并行技术的发展，我们的方法不会额外增加很多的训练时间。
% our method improves the performance significantly on several main SSL benchmarks. 

In summary, the key contributions of this work include the following:
\begin{itemize}
    \item We propose to tackle the confirmation bias problem, which is prevalent in self-training SSL methods and can lead to performance bottlenecks. The proposed CLS combines the advantages of negative learning and co-training to cope with this issue.
    \item While the previous SSL approaches mainly adopt threshold truncation to remove the influence of low-confidence pseudo labels; we propose a soft re-weighting method to quantify the quality of artificial labels and enhance the performance by improving label utilization.
    \item Comprehensive experiments demonstrate that CLS surpasses its SSL counterparts by significant margins on commonly used benchmark datasets CIFAR-10 and CIFAR-100.
\end{itemize}

%-------------------------------------------------------------------------------------------------------------
\section{Method}
\begin{figure*}
\centering
\includegraphics[width=0.842\textwidth]{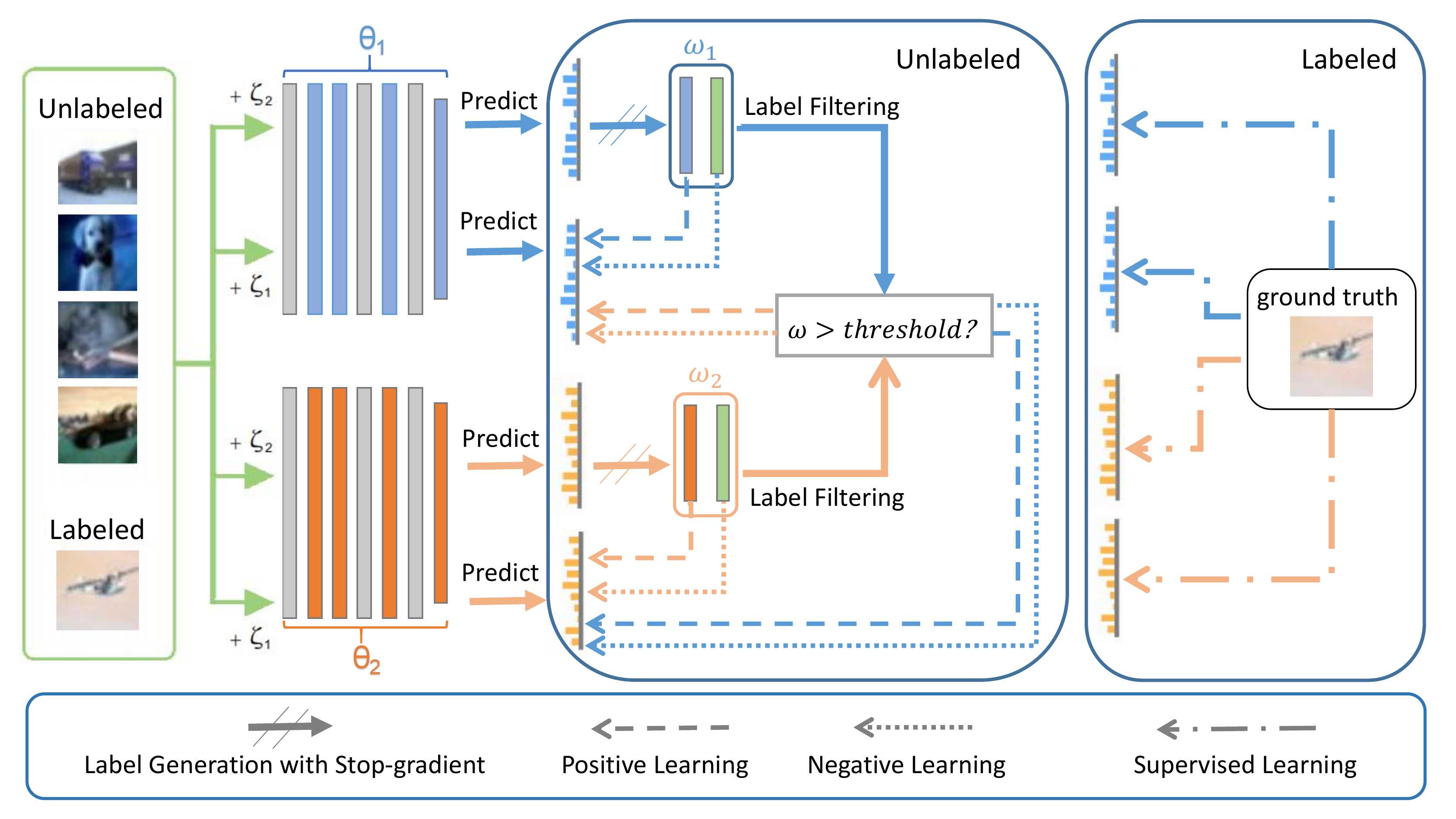}
\caption{
Structure overview of the CLS framework. 
The figure depicts a training batch with one labeled sample and four unlabeled samples. 
Before being fed into the two separate models, each sample will undergo both weak ($+ \zeta_2$) and strong ($+ \zeta_1$) augmentations. 
The prediction probability distribution for the weakly-augmented version is used to generate pseudo and complementary labels with corresponding sample weights, which are utilized to supervise the prediction on the strongly-augmented version via the positive and negative cross-entropy loss, respectively.
When exchanging knowledge between the two models, only those labels with sample weights above a threshold are considered.
As for the labeled sample, each model is trained to make its predictions match the ground-truth label.
}
\label{fig:algo-structure}
\end{figure*}

In this section, we introduce the intuition and details of CLS, which aims to overcome the confirmation bias problem. 
The complete structure of the method is presented in Figure~\ref{fig:algo-structure}.

\subsection{Notation}
For a $C$-class classification problem, let $\mathcal{X}_l=\{(x_i,y_i)\}^{N_l}_{i=1}$ denote the labeled dataset with $N_l$ samples and $\mathcal{X}_u=\{(x_i)\}^{N_u}_{i=1}$ denote the unlabeled dataset with $N_u$ samples. 
The difference between the two datasets is that the ground-truth label $y_i \in \{1,\ldots, C\}$ of input $x_i$ in $\mathcal{X}_l$ is available, whereas the corresponding labels in $\mathcal{X}_u$ are absent. 
The goal of an SSL algorithm is to optimize a classifier $f:X\rightarrow Y$, which predicts the probability distribution that the input belongs to different classes. 
The parameterization of the classifier is denoted by $\theta$, which is optimized by gradient descent:
\begin{equation}
    \theta_t \leftarrow \theta_{t-1} - \alpha\bigtriangledown_{\theta_{t-1}}L(\mathcal{X}_l\cup \mathcal{X}_u; \theta_{t-1}),
\end{equation}
where $t\in\{1, ..., T\}$ is the iteration index, $\alpha$ represents the learning rate and $L(\mathcal{X}_l\cup \mathcal{X}_u; \theta_{t-1})$ is a loss function to be specified.

%-------------------------------------------------------------------------------------------------------------
\subsection{Pseudo Label}
In order to boost the classifier's performance, a common way is to generate pseudo labels by self-labeling unlabeled samples, which are then incorporated with  $\mathcal{X}_l$ to retrain the classifier. 
Such two procedures are normally run in an iterative manner. 
There are two major forms of pseudo labeling, i.e., hard labeling and soft labeling. 
Hard labeling~\cite{lee2013pseudo} methods select the entries with the maximum probability as pseudo labels:
\begin{equation}
\label{eq:pl}
    \widetilde{y}_i=\mathop{\arg\max}\limits_{j\in \{1, \ldots,  C\}}f(x_i; \theta)_j,
\end{equation}
where $f(x_i; \theta)\in\mathbb{R}^C$ represents the predicted probability distribution of input $x_i$, and $f(x_i; \theta)_j$ denotes the corresponding probability of class $j$. 
% For the multi-label case, 
Another approach to creating hard pseudo labels is to introduce a confidence threshold $\gamma\in[0, 1]$ for truncation~\cite{rizve2021in,sohn2020fixmatch}: 
\begin{equation}
\label{hard_label_2}
    \widetilde{y}_{ij}=\mathbbm{1}\left(f\left(x_i; \theta\right)_j\geq\gamma\right), \ \text{for} \ j=1,2,\ldots, C,
\end{equation}
where $\widetilde{y}_{ij}$ denotes the $j$-th entry of the refined multi-hot label for input $x_i$, and  $\mathbbm{1}(\cdot)$ is the indicator function that outputs 1 if the inside condition is satisfied and 0 otherwise. 

Different from hard labeling, which is generally non-differentiable, soft labeling applies a sharpening function~\cite{berthelot2019mixmatch,berthelot2019remixmatch, xie2020unsupervised} to enhance high-confident prediction while demoting low-confident ones. 
An example to refine the prediction distribution is given as follows:
\begin{equation}
    \widetilde{y}_{ij}=\frac{\left(f\left(x_i; \theta\right)_j\right)^{\frac{1}{\epsilon}}}{\sum^C_{j^{\prime}=1}\left(f\left(x_i; \theta\right)_{j^{\prime}}\right)^{\frac{1}{\epsilon}}}, \text{for} \ j=1,2,\ldots, C,
\end{equation}
where $\epsilon$ is a hyper-parameter that determines the ``softness'' of soft labeling. Note that as the temperature decreases, the soft pseudo labels become sharper, especially when $\epsilon\rightarrow0$, the soft pseudo labels degrade to the hard pseudo labels.
% as Eq. (\ref{eq:pl}). 

% In conjuction with normalization~\cite{xie2016unsupervised, meng2020text}, soft labeling can counter the imbalance of pseudo labels:
% \begin{equation}
%     \widetilde{y}_{ij}=\frac{\left(f\left(x_i; \theta\right)_j\right)^{\frac{1}{\tau}}/f_{j}}{\sum^C_{j^{\prime}=1}\left[\left(f\left(x_i; \theta\right)_{j^{\prime}}\right)^{\frac{1}{\tau}}/f_{j^{\prime}}\right]}, 
% \end{equation}
% where $f_j = \sum_{i^{\prime}} \left(f\left(x_{i^{\prime}}; \theta\right)_j\right)^{\frac{1}{\tau}}$.

\subsection{Complementary Label}
Although pseudo labels can be viewed as a form of entropy minimization that moves decision boundaries to low-density regions~\cite{grandvalet2004semi, lee2013pseudo}, 
inaccurate labels' lack of calibration can lead to severe performance degradation due to the confirmation bias~\cite{kim2019nlnl,rizve2021in}. 
% inaccurate labels due to lack of calibration can result in severe performance degradation
To remedy this, CLS generates not only pseudo labels to predict what category the current input most likely belongs to, but also complementary labels to indicate what category it would not belong to. 
Complementary labels allow for negative learning~\cite{kim2019nlnl}, which aims to prevent the model from overfitting to noisy data and accelerate model training. 
In contrast to hard pseudo labels, complementary labels corresponding to the low-confidence predictions can be obtained as follows:
\begin{equation}
\label{eq:nl}
    \overline{y}_i=\mathop{\arg\min}\limits_{j\in \{1, \ldots,  C\}}f(x_i; \theta)_j.
\end{equation}
% by replacing the operator $\mathop{\arg\max}$ with $\mathop{\arg\min}$ in Eq. (\ref{hard_label_1})  or $\geq$  with $\leq$ in  Eq. (\ref{hard_label_2}).
% as follows:
% \begin{equation}
%     \overline{y}_{ij}=\mathbbm{1}\left(f\left(x_i; \theta\right)_j\leq\gamma_2\right), \text{for} \ j=1,2,\ldots, C,
% \end{equation}
% where $\gamma_2$ is a threshold hyper-parameter to be decided. 

%-------------------------------------------------------------------------------------------------------------
\subsection{Cross Labeling Supervision}
As discussed in Sec.\ref{sec:intro}, learning from artificial labels generated by the classifier itself can suffer from the confirmation bias, and we propose two countermeasures in CLS. 

\subsubsection{Weighted Labeling}
The first is to use sample re-weighting to tackle erroneous artificial labels. 
The intuition behind this is that the output softmax probability distributions with low entropy are more likely to produce accurate artificial labels. 
For example, a prediction of distribution $[0.01, 0.99]$ is more promising to infer the ground-truth label than a prediction of $[0.45, 0.55]$. Hence, it is highly desired to up-weight the more confident predictions and down-weight the less confident ones. To this end, we propose to assign adaptive sample weights to the artificial labels as follows:
\begin{equation}
    w_i = 1- \frac{H\left(f\left(x_i;\theta\right)\right)}{\log(C)},
\end{equation}
where $H\left(f\left(x_i;\theta\right)\right)=-\sum^C_{j=1}f\left(x_i; \theta\right)_j\log\left(f\left(x_i; \theta\right)_j\right)$ is the entropy of $f(x_i; \theta)$ bounded in the interval $[0, \log(C)]$. 
Notice that $w_i\in[0, 1]$ can reflect statistics regarding the confidence of artificial labels.
Specifically, $w_i\rightarrow0$  is equivalent to the uniform distribution (uncertain labels), whereas $w_i\rightarrow1$ implies the one-hot distribution (deterministic labels).

Recall that hard labeling has become a typical configuration in SSL research due to its simplicity, generality, and ease of implementation. 
In this paper, we propose the re-weighting  mechanism to maintain these advantages while mitigating the effects of mislabeling to enhance SSL performance. 
In particular,  two modified cross-entropy loss functions are utilized as follows:
\begin{flalign}
    && \mathcal{L}_{P}(\theta, x_i, w_i, \widetilde{y}_i)  & = -w_i\sum^C_{j=1}\widetilde{y}_{ij}\log {f(x_i; \theta)_j}, & \label{loss_eq1}\\
    && \mathcal{L}_{N}(\theta, x_i, w_i, \overline{y}_i)  & = -w_i\sum^C_{j=1}\overline{y}_{ij}\log \left(1-{f\left(x_i; \theta\right)_j}\right), & \label{loss_eq2}
\end{flalign}
where $\widetilde{y}_i /\overline{y}_i$ is a given pseudo/complementary label with a discounted sample weight of $w_i$, and 
% of a unlabeled sample $x_i \in \mathcal{X}_u$, respectively,  
$\widetilde{y}_{ij} / \overline{y}_{ij}$ represents the $j$-th element of corresponding one-hot vectors.
%, and $w_i$ is calculated during label generation.

% Furthermore, the ensemble of lightweight 1-step models adds little computational overhead as they are trained together efficiently in parallel across all time steps.

%---------------------------------------------------------------------------------------------------------
\subsubsection{Cross Labeling}
\label{sec:cross}
Inspired by co-training~\cite{blum1998combining, han2018co,ke2019dual}, we fuse knowledge from two collaborative models to alleviate the confirmation bias.
To put this idea into practice, CLS trains two independent models simultaneously, adding little computational overhead due to parallel training at all time steps.
% To put this idea into practice, CLS simultaneously trains two independent models, which add little computational overhead due to  parallel training across all time steps.
However, the predictions of the two models may be inconsistent, and directly enforcing the consistency constraint between their output will mislead them to collapse into each other even though their initial states are different.
Recall that diversity or different views play a non-trivial role in the success of co-training~\cite{blum1998combining}.
In order to maintain the independence and difference between the two models, only high-confidence artificial labels are exchanged so that the supervised signals obtained from each other are always beneficial.
Specifically, we introduce a weight threshold $\tau$ to filter out artificial labels whose sample weight is negligible.
% Fixmatch~\cite{sohn2020fixmatch} has a similar threshold, which is used to select pseudo labels for 
% Following FixMatch~\cite{sohn2020fixmatch}, each network also leverages the targets from the weak-augmented samples to guide itself. 

\subsubsection{Training Procedure}
We briefly elaborate on the training process of CLS. It contains two independent models whose network weights are denoted by $\theta^1$ and $\theta^2$, respectively.
Similar to FixMatch~\cite{sohn2020fixmatch}, each network generates the targets from the weak-augmented samples:
\begin{flalign}
&& \widetilde{y}_i^b &=\mathop{\arg\max}\limits_{j\in \{1, \ldots,  C\}}f(x_i+ \zeta_2; \theta^b)_j, & \label{eq:self_pl} \\
&& \overline{y}_i^b &=\mathop{\arg\min}\limits_{j\in \{1, \ldots,  C\}}f(x_i+ \zeta_2; \theta^b)_j,  & \label{eq:self_nl} \\
&& w_i^b &=1- \frac{H\left(f\left(x_i+ \zeta_2;\theta^b\right)\right)}{\log(C)}, & \label{eq:self_w}
\end{flalign}
where $\zeta_2$ represents a weak augmentation strategy, and $b\in\{1,2\}$ is an identifier to indicate the corresponding network. 
With the generated artificial labels and corresponding sample weights, for a mini-batch $\mathcal{X}_u^B \subseteq  \mathcal{X}_u$,  we define the self-labeling loss function as 
\begin{flalign}
 \label{eq:self_l}
    L_{self}^b(\mathcal{X}_u^B) = \frac{1}{B}\sum_{x_i \in \mathcal{X}_u^B} \mathcal{L}_{P}(\theta^b, x_i&+\zeta_1, w_i^b, \widetilde{y}_i^b) \nonumber \\
    + \mathcal{L}_{N}(\theta^b, x_i&+\zeta_1, w_i^b, \overline{y}_i^b), 
\end{flalign}
where $\zeta_1$ denotes the strong augmentation strategy, and $B$ is the mini-batch size. 

As analyzed in Sec.~\ref{sec:cross}, supervised signals from the weakly-augmented samples may not be adequate to calibrate the confirmation bias in self-training.
To address this issue, we utilize cross labeling and define the co-labeling loss function concerning $\theta^1$ as
\begin{flalign}
 \label{eq:cross_l}
    L_{co}^1(\mathcal{X}_u^B) &= \frac{1}{B}\sum_{x_i \in \mathcal{X}_u^B} \mathbbm{1}(w_i^2 > \tau) \mathcal{L}_{P}(\theta^1, x_i+\zeta_1, w_i^{2},\widetilde{y}_i^{2}) \nonumber \\
    &+ \mathbbm{1}(w_i^2 > \tau) \mathcal{L}_{N}(\theta^1, x_i+\zeta_1, w_i^2,\overline{y}_i^2),
\end{flalign}
where $\tau\in [0, 1]$ is the weight threshold, and a larger value usually means that less but higher quality information is exchanged.
In particular, cross labeling would fail when setting $\tau=1$ because $w_i^b$ ranges from 0 to 1, while setting $\tau=0$ may risk the model collapsing into each other. 
$L_{co}^2(\mathcal{X}_u^B)$ can be calculated in a similar way to Eq. (\ref{eq:cross_l}).

As for a labeled mini-batch $\mathcal{X}_l^B \subseteq  \mathcal{X}_l$, both models are trained via a standard supervised classification as 
\begin{equation}
    L_{sup}^b(\mathcal{X}_l^B) = \frac{1}{B}\sum_{(x_i, y_i) \in \mathcal{X}_l^B} {\sum^C_{j=1}{y}_{ij}\log {f(x_i; \theta^b)_j}}.
\end{equation}
% where ${y}_{ij}$ indicates the $j$-th element of ground truth $y_i$.

Following FixMatch~\cite{sohn2020fixmatch}, we consider different relative sizes of $\mathcal{X}_u$ and $\mathcal{X}_l$, which are denoted by $\mu$. Given $\mathcal{X}_l^B$ and  $\mathcal{X}_u^{\mu B}$, network $\theta^b$ ($b\in\{1,2\}$) is optimized with the mixed loss in the form of
\begin{equation}
\label{eq:mixed}
    L^b = L_{sup}^b(\mathcal{X}_l^B)+\lambda_1 L_{self}^b(\mathcal{X}_u^{\mu B}) + \lambda_2 L_{co}^b(\mathcal{X}_u^{\mu B}),
\end{equation}
where $\lambda_1$ and $\lambda_2$ are the trade-off coefficients between various losses.
The complete algorithm for CLS is presented in algorithm~\ref{algo:our-method}.

\begin{algorithm}[t]
\caption{CLS Algorithm}
\label{algo:our-method}
\SetAlgoLined
\KwIn{$\mathcal{X}_l$: labeled dataset; $\mathcal{X}_u$: unlabeled dataset;\quad $\mu$: unlabeled data ratio; $\tau$: weight threshold; $\lambda_1$/$\lambda_2$: self/co-labeling loss coefficient; \quad
$\zeta_1$/$\zeta_2$: strong/weak augmentation strategy.}
Initialize $\theta^1_0$ and $\theta^2_0$ with different random seeds; \\
\For{$t=1,\cdots,T$}{
    Sample a mini-batch of size $B$ from $\mathcal{X}_l$: $\mathcal{X}_l^B$;   \\
    Sample a mini-batch of size $\mu B$ from $\mathcal{X}_u$: $\mathcal{X}_u^{\mu B}$; \\
    \tcp{Label generation}
    \For{$b=1, 2$}{
        Pseudo labels $\leftarrow$ Eq. (\ref{eq:self_pl}) on $\mathcal{X}_u^{\mu B}$; \\
        Complementary labels $\leftarrow$ Eq. (\ref{eq:self_nl}) on $\mathcal{X}_u^{\mu B}$; \\
        Sample weights $\leftarrow$ Eq. (\ref{eq:self_w}) on $\mathcal{X}_u^{\mu B}$;
    }
    \tcp{Loss calculation}
    \For{$b=1, 2$}{
        $L^b \leftarrow$ Eq. (\ref{eq:mixed}) on $\mathcal{X}_l^B \cup \mathcal{X}_u^{\mu B}$;
    }
    \tcp{Update $\theta_1$ and $\theta_2$ in parallel}
    $\theta_t^1 \leftarrow \theta_{t-1}^1 - \alpha\bigtriangledown_{\theta_{t-1}^1}L^1$;\\
    $\theta_t^2 \leftarrow \theta_{t-1}^2 - \alpha\bigtriangledown_{\theta_{t-1}^2}L^2$;\\
}
\KwOut{$\theta_T^1$ or $\theta_T^2$ for prediction.}
\end{algorithm}
%---------------------------------------------------------------------------------------------------------
\section{Experiments}
Our experiments aim to answer the following questions:
\begin{itemize}
    \item \textbf{RQ1}: How does CLS perform in standard benchmarks compared to prior state-of-the-art SSL algorithms?
    \item \textbf{RQ2}: What is the effect of each component in CLS?
    \item \textbf{RQ3}: How does CLS mitigate the confirmation bias?
    \item \textbf{RQ4}: How do different hyperparameters affect CLS?
\end{itemize}
% Next, we describe the experiments performed to evaluate our method.

\subsection{Experimental Settings}
\subsubsection{Datasets}
The benchmark datasets we used are CIFAR-10, CIFAR-100~\cite{krizhevsky2009learning}, which both contain 50K training images and 10K test images of size 32 × 32 with respect to 10 and 100 class categories, respectively. Following the partition protocols of Fixmatch~\cite{sohn2020fixmatch}, we experiment with three sizes of labeled images on those two datasets. To be specific, we divide the training set of CIFAR-10 into two groups, with 40, 250, and 4K samples randomly selected as the labeled set  and the rest as the unlabeled set. CIFAR-100 is similarly divided, corresponding to 400, 2.5K, and 10K labeled samples. The prediction accuracy on the test set is adopted as the evaluation metric.

\subsubsection{Baselines}
We compare CLS against four groups of baselines. 
The vanilla baseline is supervised learning (SL) with data augmentation methods, e.g., RandAugment (RA)~\cite{cubuk2020randaugment}.
This baseline is set to ensure that SL has a relatively fair comparison with those state-of-the-art SSL methods that leverage strong data augmentation methods.
Note that the RA method is also applied to UPS~\cite{rizve2021in}, UDA~\cite{xie2020unsupervised}, and all hybrid methods.
The second group of baselines consists of two self-training methods that generate pseudo labels for the unlabeled set, where high-confidence labels are selected to expand the labeled set.
Specifically, Pseudo-Label~\cite{lee2013pseudo} generates pseudo labels without using data augmentation methods, while UPS~\cite{rizve2021in} creates additional complementary labels for the low-confidence samples in the unlabeled dataset.
The third group contains consistency regularization methods, where $\Pi$-model~\cite{rasmus2015semi} and Mean Teacher~\cite{tarvainen2017mean} are two classic benchmarks while UDA~\cite{xie2020unsupervised} represents the state-of-the-art consistency training method.
Similar to CLS, recent hybrid methods incorporate self-training with consistency training. Three state-of-the-art methods are selected for comparison, i.e., MixMatch~\cite{berthelot2019mixmatch}, ReMixMatch~\cite{berthelot2019remixmatch}, and FixMatch~\cite{sohn2020fixmatch}.

\subsubsection{Implementation Details}
We implement our method based on the PyTorch~\cite{paszke2019pytorch} framework, \emph{i.e.}, PyTorch 1.3. In our experiments, all baselines share the same backbone and dataset partitioning. For CIFAR-10, the backbone architecture adopted is a WideResNet-28-2~\cite{zagoruyko2016wide} with 1.45 million parameters, while a WideResNet-28-8~\cite{zagoruyko2016wide} with 23.40 million parameters is used for CIFAR-100. This setting is commonly used by previous works~\cite{rizve2021in,sohn2020fixmatch, berthelot2019mixmatch}. We utilize the SGD optimizer~\cite{bottou2012stochastic} in conjunction with Nesterov momentum~\cite{tang2018adaptive} for network training. Following FixMatch~\cite{sohn2020fixmatch}, the cosine learning rate decay schedule is used for learning rate adjustment. 
In all of our experiments, weak augmentation is realized with a standard flip-and-shift augmentation strategy, while the strong augmentation strategy is RandAugment (RA)~\cite{cubuk2020randaugment}. All experiments are trained on NVIDIA Tesla V100 GPUs, and the default hyper-parameter configuration of CLS is \{$\alpha =0.03, \mu =8, B=64, T=300, \tau=0.85, \lambda_1=2, \lambda_2=1$\}.
Refer to Appendix for more details.
% We use SGD optimizer with an initial learning rate of 0.03 and cosine annealing~\cite{loshchilov2016sgdr} for learning rate decay. 
% The momentum is fixed as 0.9 and the weight decay is set to 0.0005. We employ a poly learning rate policy where the initial learning rate is multiplied by 1. We set the confidence thresholds N= 0.7 and N = 0.05 for all experiments. Furthermore, for the uncertainty thresholds we use A = 0.05 and B = 0.005 for all experiments.

\begin{table*}[t]
\caption{Classification accuracy (\%) of CLS and various baselines over 4 runs on 2 standard benchmarks, the higher, the better.}
\label{tab:comparison-table}
\begin{tabular}{l|l|lll|lll}
\toprule[2pt]
\multirow{2}{*}{\textbf{Category}}                                                   & \multirow{2}{*}{\textbf{Method}}          & \multicolumn{3}{c|}{\textbf{CIFAR-10}}                                                                               & \multicolumn{3}{c}{\textbf{CIFAR-100}}                                                                               \\
                                                                                     &                                           & \multicolumn{1}{c}{40 labels}            & \multicolumn{1}{c}{250 labels}           & \multicolumn{1}{c|}{4k labels} & \multicolumn{1}{c}{400 labels}           & \multicolumn{1}{c}{2.5k labels}          & \multicolumn{1}{c}{10k labels} \\ 
\midrule[1pt]
Vanilla baseline                                                                             & SL w/ RA~\cite{cubuk2020randaugment}                 & \multicolumn{1}{l|}{35.96$\pm$0.88}          & \multicolumn{1}{l|}{60.78$\pm$0.65}          & 87.36$\pm$0.32                     & \multicolumn{1}{l|}{20.43$\pm$0.22}          & \multicolumn{1}{l|}{46.93$\pm$0.47}          & 68.42$\pm$0.33                     \\ \hline
\multirow{2}{*}{Self-Training}     & Pseudo-Label~\cite{lee2013pseudo}              & \multicolumn{1}{l|}{-}                   & \multicolumn{1}{l|}{51.31$\pm$0.66}          & 84.62$\pm$0.38                     & \multicolumn{1}{l|}{-}                   & \multicolumn{1}{l|}{43.26$\pm$0.43}          & 65.78$\pm$0.29                     \\ \cline{3-8} 
                                                                                     & UPS~\cite{rizve2021in}                                       & \multicolumn{1}{l|}{-}                   & \multicolumn{1}{l|}{78.39$\pm$0.88}          & 89.59$\pm$0.64                     & \multicolumn{1}{l|}{-}                   & \multicolumn{1}{l|}{52.34$\pm$0.45}          & 68.77$\pm$0.25                     \\ \hline
\multirow{3}{*}{\begin{tabular}[c]{@{}l@{}}Consistency\\ Training\end{tabular}} & $\Pi$-model~\cite{rasmus2015semi}                                   & \multicolumn{1}{l|}{-}                   & \multicolumn{1}{l|}{46.24$\pm$5.48}          & 86.41$\pm$0.56                     & \multicolumn{1}{l|}{-}                   & \multicolumn{1}{l|}{43.85$\pm$0.68}          & 63.53$\pm$0.16                     \\
                                                                                     & Mean Teacher~\cite{tarvainen2017mean} & \multicolumn{1}{l|}{-}                   & \multicolumn{1}{l|}{69.44$\pm$3.11}          & 91.22$\pm$0.21                     & \multicolumn{1}{l|}{-}                   & \multicolumn{1}{l|}{47.19$\pm$0.62}          & 66.21$\pm$0.33                     \\ 
                                                                                     & UDA~\cite{xie2020unsupervised}                    & \multicolumn{1}{l|}{71.63$\pm$6.72}          & \multicolumn{1}{l|}{91.10$\pm$1.15}          & 95.08$\pm$0.22                     & \multicolumn{1}{l|}{41.24$\pm$0.92}          & \multicolumn{1}{l|}{66.89$\pm$0.32}          & 75.23$\pm$0.45                     \\
                                                                                     \hline
\multirow{3}{*}{Hybrid methods}                                                      & MixMatch~\cite{berthelot2019mixmatch}                                  & \multicolumn{1}{l|}{54.65$\pm$9.78}          & \multicolumn{1}{l|}{89.21$\pm$1.22}          & 93.66$\pm$0.18                     & \multicolumn{1}{l|}{33.43$\pm$2.14}          & \multicolumn{1}{l|}{61.14$\pm$0.67}          & 72.17$\pm$0.42                     \\
                                                                                     & ReMixMatch~\cite{berthelot2019remixmatch}                                & \multicolumn{1}{l|}{81.94$\pm$7.63}          & \multicolumn{1}{l|}{94.48$\pm$0.36}          & 95.26$\pm$0.14                     & \multicolumn{1}{l|}{\textbf{\underline{54.88}}$\pm$2.33} & \multicolumn{1}{l|}{\textbf{\underline{73.37}}$\pm$0.41} & 76.88$\pm$0.63                     \\
                                                                                     & FixMatch~\cite{sohn2020fixmatch}             & \multicolumn{1}{l|}{\textbf{\underline{88.29}}$\pm$3.44} & \multicolumn{1}{l|}{\textbf{\underline{94.85}}$\pm$0.76} & \textbf{\underline{95.73}}$\pm$0.15            & \multicolumn{1}{l|}{51.81$\pm$2.16}          & \multicolumn{1}{l|}{72.73$\pm$0.66}          & \textbf{\underline{77.33}}$\pm$0.24            \\ \hline
Hybrid methods                                                                       & CLS                                       & \multicolumn{1}{l|}{\textbf{91.82}$\pm$1.77} & \multicolumn{1}{l|}{\textbf{95.55}$\pm$0.33} & \textbf{96.28}$\pm$0.06            & \multicolumn{1}{l|}{\textbf{55.91}$\pm$1.12} & \multicolumn{1}{l|}{\textbf{74.06}$\pm$0.28} & \textbf{79.27}$\pm$0.16            \\ 
\bottomrule[2pt]
\end{tabular}
\end{table*}

\subsection{Comparison with Baselines}
Table~\ref{tab:comparison-table} shows  the results on CIFAR-10 and CIFAR-100 with different sizes of the labeled dataset.
We report the averaged test accuracy and corresponding standard deviation over four runs.
All hybrid methods consistently outperform the vanilla baseline by substantial margins across all settings, showing the promise of SSL methods.
It is worth mentioning that the vanilla baseline can perform on par with traditional SSL methods that do not use data augmentation, i.e., Pseudo-Label~\cite{lee2013pseudo}, $\Pi$-model~\cite{rasmus2015semi}, and Mean Teacher~\cite{tarvainen2017mean}.
Such results prove the necessity and effectiveness of advanced data augmentation methods.
Furthermore, CLS achieves state-of-the-art performances in all settings, which demonstrates the validity of our method, especially in highly label-scarce settings.
Specifically, we surpass the vanilla baseline by considerable margins of 55.86\% and 35.48\% accuracy on CIFAR-10 with 40 labels and CIFAR-100 with 400 labels.
More importantly, although CLS can be viewed as an extension of FixMatch~\cite{sohn2020fixmatch},  stable performance gains achieved by CLS on both datasets suggest that the modifications to FixMatch are effective.

% Furthermore, evident 9.7\% gains are achieved on tasks. 

\begin{table*}[t]
\caption{Classification accuracy (\%) of several variants of CLS on CIFAR-10 and CIFAR-100 over 4 runs. 
% All models follow the default setting and are tested using the same codebase.
}
\label{tab:ablation-table}
\begin{tabular}{l|lll|lll}
\toprule[2pt]
\multirow{2}{*}{\textbf{Method}}     & \multicolumn{3}{c|}{\textbf{CIFAR-10}}                                                                 & \multicolumn{3}{c}{\textbf{CIFAR-100}}                                                                 \\
                            & \multicolumn{1}{c}{40 labels}          & \multicolumn{1}{c}{250 labels}         & \multicolumn{1}{c|}{4k labels} & \multicolumn{1}{c}{400 labels}         & \multicolumn{1}{c}{2.5k labels}        & \multicolumn{1}{c}{10k labels} \\
                            \midrule[1pt]
FixMatch                    & \multicolumn{1}{l|}{$88.29\pm3.44$} & \multicolumn{1}{l|}{$94.85\pm0.76$} & $95.73\pm0.15$                & \multicolumn{1}{l|}{$51.81\pm2.16$} & \multicolumn{1}{l|}{$72.73\pm0.66$} & $77.33\pm0.24$                \\ \hline
FixMatch w/ NL               & \multicolumn{1}{l|}{$89.58\pm2.11$} & \multicolumn{1}{l|}{$94.91\pm0.87$} & $95.89\pm0.07$                & \multicolumn{1}{l|}{$52.76\pm1.61$} & \multicolumn{1}{l|}{$73.21\pm0.46$}  & $77.45\pm0.54$                \\
FixMatch w/ RW      & \multicolumn{1}{l|}{$89.87\pm1.98$}           & \multicolumn{1}{l|}{$94.89\pm1.06$} & $95.82\pm0.11$                & \multicolumn{1}{l|}{$52.43\pm2.87$} & \multicolumn{1}{l|}{$73.37\pm0.43$} & $77.75\pm0.32$                \\
FixMatch w/ (NL \& RW) & \multicolumn{1}{l|}{$90.24\pm1.87$ }           & \multicolumn{1}{l|}{\textbf{\underline{94.98}} $\pm$ 0.65} & $96.02\pm0.29$              & \multicolumn{1}{l|}{$53.42\pm1.33$} & \multicolumn{1}{l|}{\textbf{\underline{73.49}} $\pm$ 0.35} & $78.31\pm0.18$                \\ \hline
CLS w/o NL          & \multicolumn{1}{l|}{$90.61\pm2.64$}           & \multicolumn{1}{l|}{$94.87\pm1.11$} & $96.03\pm0.09$              & \multicolumn{1}{l|}{$52.87\pm1.31$} & \multicolumn{1}{l|}{$72.78\pm0.39$} & $78.55\pm0.25$               \\
CLS w/o RW & \multicolumn{1}{l|}{\textbf{\underline{90.89}} $\pm$ 2.57} & \multicolumn{1}{l|}{$94.91\pm0.87$} & \textbf{\underline{96.05}} $\pm$ 0.07                & \multicolumn{1}{l|}{\textbf{\underline{53.59}} $\pm$ 1.95} & \multicolumn{1}{l|}{$73.34\pm0.41$} & \textbf{\underline{78.69}} $\pm$ 0.33                \\ \hline
CLS                 & \multicolumn{1}{l|}{\textbf{91.82 $\pm$ 1.77}}           & \multicolumn{1}{l|}{\textbf{95.55 $\pm$ 0.33}} & \textbf{96.28 $\pm$ 0.06 }             & \multicolumn{1}{l|}{\textbf{55.91 $\pm$ 1.12}} & \multicolumn{1}{l|}{\textbf{74.06 $\pm$ 0.28}} & \textbf{79.27 $\pm$ 0.16}                \\ 
\bottomrule[2pt]
\end{tabular}
\end{table*}

\begin{figure}[t]
\centering
\includegraphics[width=0.49\textwidth]{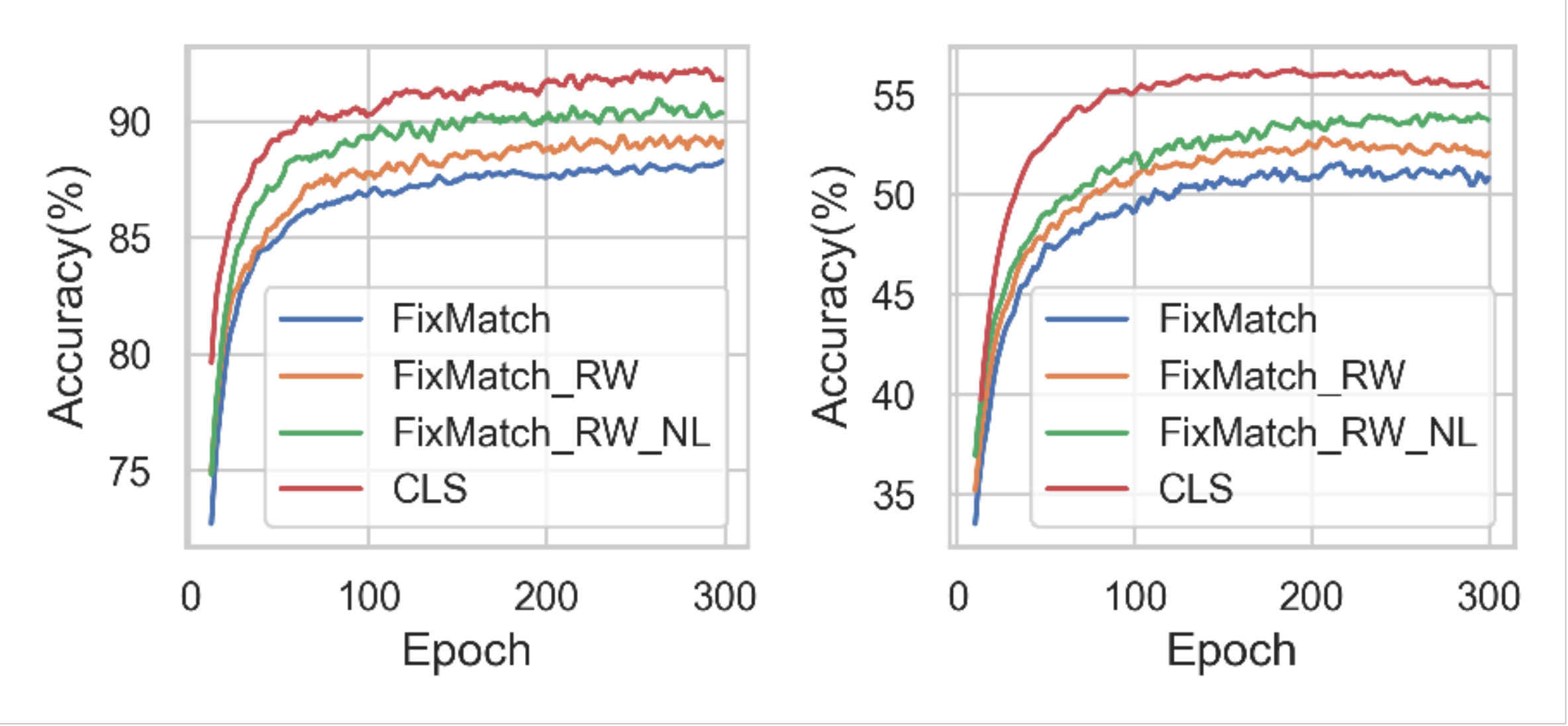}
\caption{
The classification accuracy of FixMatch and three variants of CLS. \textbf{Left:} Test accuracy on CIFAR-10 with 40 labels. \textbf{Right:} Test accuracy on CIFAR-100 with 400 labels. The gap between the four curves indicates that the introduction of each component brings a performance improvement.
}
\label{fig:curves}
\end{figure}

\subsection{Ablation Study}
Our paper suggests a  comprehensive framework for SSL consisting of the following essential ingredients: (1) using sample re-weighting mechanism, short for \textbf{RW}; (2) generating extra complementary labels for negative learning, abbreviated as \textbf{NL}; (3) exchanging artificial labels for cross supervision. 
To investigate the strength of each component in CLS, we conducted an analysis that reveals the performance differences when using different combinations of these components during training.
Variants include modifications to FixMatch and the removal of \textbf{NL} or \textbf{RW} from the CLS.
All variants use the default configuration of CLS, except for changes in specific components.
% Experiments include adding modifications to FixMatch and deleting either \textbf{NL} or \textbf{RW} from CLS. 
% The default parameter configuration is used for all experiments.
% All variants use the default configuration of CLS for the remaining hyper-parameters except for the change of specific components.
% Table ~\ref{tab:ablation-table} shows all experiments conducted for the analysis, following the experimental settings of Table ~\ref{tab:comparison-table} with three different labeled size on CIFAR-10 and CIFAR-100.
Table~\ref{tab:ablation-table} presents the results of all experiments performed for the analysis.
% following the experimental setup in Table ~\ref{tab:comparison-table} with labeled datasets of three different sizes on CIFAR-10 and CIFAR-100.

As shown in Table~\ref{tab:ablation-table}, compared to FixMatch, all variants achieve performance improvements, demonstrating the positive effect of each component.
It is noteworthy that significant performance gains are shown when there are fewer labeled images, especially in the two cases of CIFAR-10 with 40 labels and CIFAR-100 with 400 labels, where only 4 labels are assigned to each class on both datasets.
Specifically, CLS improved the average accuracy from 88.29\% to 91.82\% on CIFAR-10 with 40 labels and from 51.81\% to 55.91\% on CIFAR-100 with 400 labels.
To explore whether the efficacy of different components can be superimposed, we visualized the learning curves of  four variants in label-scarce settings.  
As shown in Figure~\ref{fig:curves}, the increase in accuracy suggests that the three components of CLS complement each other, and their combination can yield better accuracy than omitting one or more elements.

\begin{figure}[t]
\centering
\includegraphics[width=0.49\textwidth]{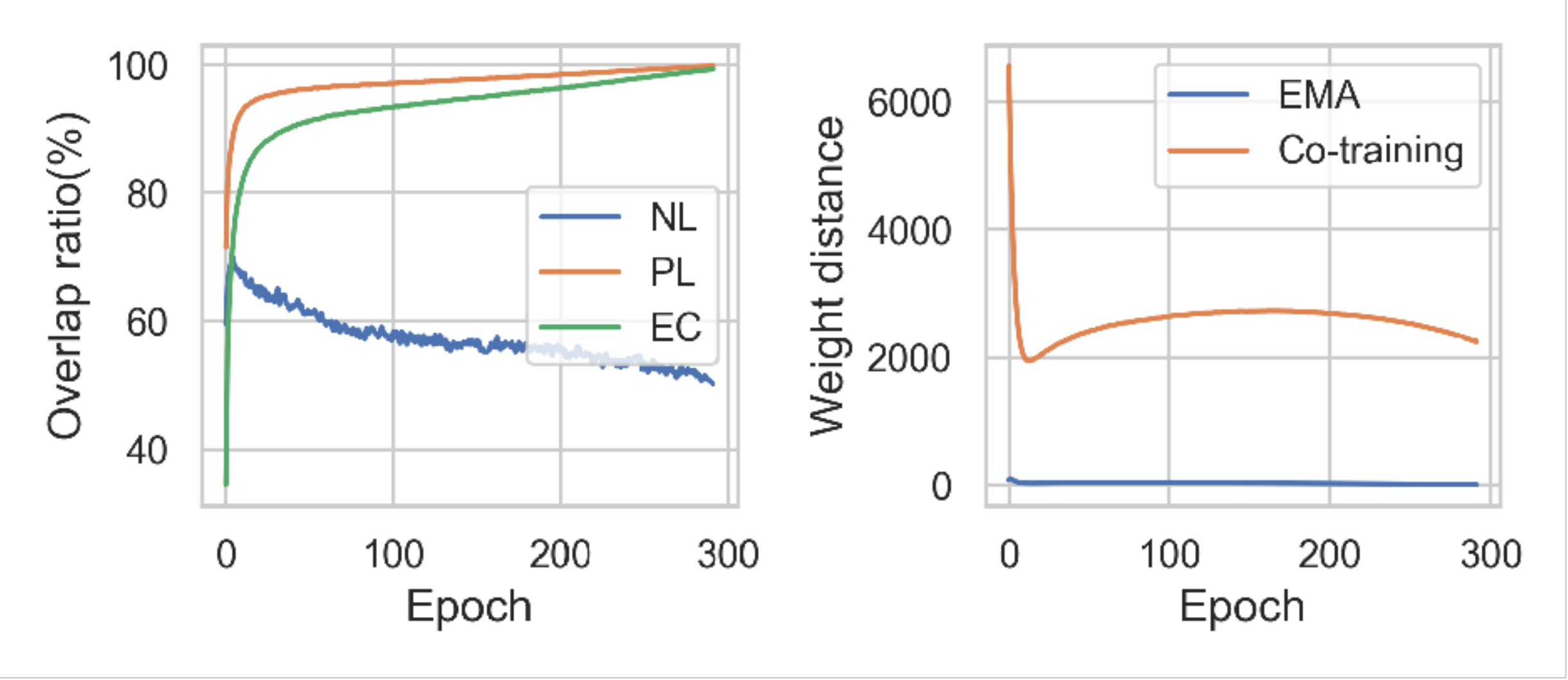}
\caption{
Illustration of the cross labeling.
\textbf{Left:} While the overlap of pseudo labels (PL) generated by the two networks and the exchange ratio (EC) of labels increases over time, the overlap of complementary labels (NL) predicted by the two networks remains at a relatively low level.
\textbf{Right:} An exponential moving average (EMA) of network $\theta^1$ has similar weights to itself, while the weights of $\theta^1$ and $\theta^2$ keep a certain distance.
% to encourage diversity.
}
\label{fig:analysis}
\end{figure}

\subsection{Analysis of Cross Labeling}
As discussed in Sec.~\ref{sec:cross}, the confirmation bias is an inevitable obstacle to self-labeling methods. For instance, using artificial labels from weakly-augmented samples as targets for strongly-augmented samples may risk overfitting inaccurate labels since the predictions for the strongly-augmented samples come from the same network. 
In addition, recent self-labeling methods generally use exponential moving average (EMA) techniques to provide more stable predictions~\cite{tarvainen2017mean}. However, the coupling effect of EMA models is likely to result in an accumulation of errors and make misclassification irreversible by enforcing the current predictions to match those of the EMA. In our work, we propose to use cross labeling, a specific variant of co-training~\cite{blum1998combining}, to address this issue. 
% Furthermore, recent self-training methods generally use the exponential moving average (EMA) technique to provide stable predictions. However, the coupling effect of an EMA model is likely to result in an accumulation of errors if prior predictions are inaccurate and makes the misclassification irreversible by continuously forcing the current prediction to match the prediction of EMA.
% This is a case of the confirmation bias, which is an inevitable obstacle to self-training. 

% To visualize how cross labeling works, we count the proportion of identical artificial labels generated by the two models and the proportion used for exchange. 
% To visualize why cross-labeling is effective, we calculated the proportion of identical artificial labels generated by the two models and the proportion used for exchange.
% We also calculated the Euclidean distance of different model weights. The results are visualized in Figure~\ref{fig:analysis}. As expected, the predictions of the two models become gradually consistent as the training proceeds, but the weights of the two still remain quite different and can provide a certain percentage of different complementary labels to complement each other's training.

To visualize why cross-labeling is effective, we calculated the proportion of identical artificial labels generated by the two models and the ratio used for exchange.
We also calculated the Euclidean distance of different model weights. The results are visualized in Figure~\ref{fig:analysis}. As expected, the predictions of the two models become gradually consistent as the training proceeds. However, the weights of the two collaborative models remain pretty different and can provide a certain percentage of inconsistent complementary labels to complement each other's training.

\begin{figure}[t]
\centering
\includegraphics[width=0.47\textwidth]{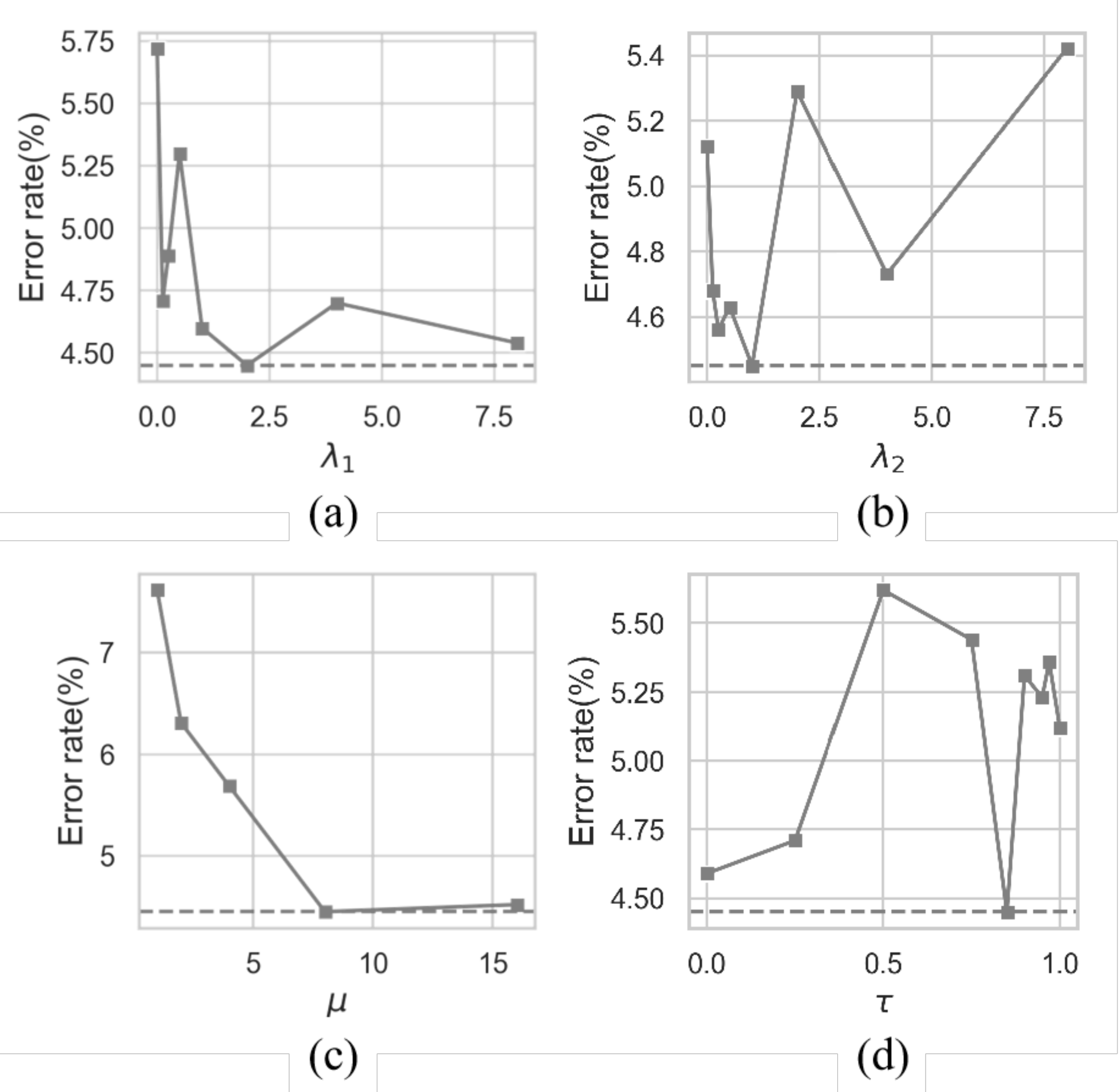}
\caption{
Test error rate on CIFAR-10 with 250 labels. (a) Varying self-labeling loss coefficient $\lambda_1$. (b) Varying co-labeling loss coefficient $\lambda_2$. (c) Varying  the ratio of unlabeled data. (d) Varying the weight threshold
for cross labeling. In each experiment, we varied one hyper-parameter, and used the default hyper-parameters for the rest. Error rate of CLS with the default setting is in dotted line.
}
\label{fig:sense}
\end{figure}

\subsection{Hyper-parameter Study}
CLS has four key hyper-parameters: the trade-off coefficients between various losses ($\lambda_1$ and $\lambda_2$), the ratio of unlabeled samples to labeled samples in each mini-batch ($\mu$), and the weight-threshold hyper-parameter $\tau$ for exchanging artificial labels between two networks. 
To assess the sensitivity of various aspects of CLS, we ran experiments on CIFAR-10 with 250 labels, varying one hyper-parameter at a time while keeping the others fixed.

\textbf{Necessity of the self-labeling loss} (Figure~\ref{fig:sense}(a)). When the self-labeling loss is given small coefficients, performance degradation can be observed. In particular, when the self-supervised loss is removed, i.e., $\lambda_1 = 0$, the error rate increases by 1.25\%, which is a remarkable gap on CIFAR-10 with 250 labels. On the other hand, in the case of $\lambda_1 \geq 1$, the performance of CLS is relatively stable.

\textbf{Effectiveness of the co-labeling loss} (Figure~\ref{fig:sense}(b) and ~\ref{fig:sense}(d)). Similar to the self-labeling loss, performance decreases with the removal of the co-labeling loss ($\lambda_2=0$ or $\tau=1$). An interesting finding is that the performance also degrades when the coefficient of the co-labeling loss is too large ($\lambda_2 \geq 2$). This is because the two networks may disagree on certain predictions, and pseudo labels provided by the other network may not be accurate. However, this property also allows the co-labeling loss to be treated as a regularization term to prevent overfitting. 

\textbf{ CLS prefers larger values of $\mu$} (Figure~\ref{fig:sense}(c)). A significant decrease in error rates can be observed with the increase of unlabeled data, which is in agreement with the conclusions in UDA~\cite{xie2020unsupervised} and FixMatch~\cite{sohn2020fixmatch}.

%---------------------------------------------------------------------------------------------------------
\section{Related Work}
% Semi-supervised learning is a mature field with a huge diversity of approaches. In this review, we focus on methods closely related to CLS.
Semi-supervised learning is a mature field with a vast diversity of approaches. In this review, we focus on methods closely related to CLS.

\subsection{Self-training Methods}
The idea of self-training originates from~\cite{mclachlan1975iterative}, which derives a paradigm that leverages the model to generate artificial labels for unlabeled data and then utilizes artificial labels to re-optimize the model itself, with the two processes alternating iteratively.
Due to its generality and simplicity, self-training has been widely used in many fields, such as image classification~\cite{xie2020self}, object detection~\cite{yang2021interactive}, etc.
Pseudo-labeling~\cite{lee2013pseudo} is a special case of self-training 
% that corresponds to generating hard labels~\cite{lee2013pseudo},
and is usually used in conjunction with confidence-based thresholds, i.e., only unlabeled samples with prediction confidence above a certain threshold are used to supplement the training set.
As described in~\cite{grandvalet2004semi}, pseudo-labeling is a form of entropy minimization that can produce better results, and it has become a standard component of the SSL algorithm pipeline.
Although pseudo-labeling has achieved excellent performance in various tasks, some studies suggested that it can suffer from the vulnerability to inaccurate pseudo-labels~\cite{arazo2020pseudo}, especially in large-sized label space.
In addition to the confidence-based threshold mechanism, another common solution to mitigate the effects of noisy labels is sample reweighting~\cite{kumar2010self}, where high-confidence samples are granted greater weights, and vice versa.
Recently, UPS~\cite{rizve2021in} introduces the concept of negative learning~\cite{kim2019nlnl} and generates complementary labels when pseudo-labels are not sufficiently confident.

\subsection{Consistency Regularization}
The idea behind consistency  regularization in SSL literature is to require the classifier to be robust to  stochastic transformations and perturbations~\cite{sajjadi2016regularization}, which was first proposed by~\cite{bachman2014learning}. 
To this end, input perturbation methods~\cite{sajjadi2016regularization} 
% apply various perturbations or noise to input samples and  
impose consistency constraints on the predictions between different augmentations of the same sample so that the decision bound can lie in the low-density region. 
On the other hand, feature perturbation methods perturb the model structure (e.g., using dropout~\cite{srivastava2014dropout}) or utilize multiple models~\cite{ke2019dual} to get multiple outputs on which consistency constraints are imposed.
To be specific, ``$\Pi$-Model''~\cite{rasmus2015semi} forces the predictions of two different augmentations of the same image to match each other.
% Compared to the independent noise used in ``$\Pi$-Model'', Virtual Adversarial Training~\cite{miyato2018virtual} is another input perturbation method that generates adversarial perturbations. 
Temporal Ensembling~\cite{laine2016temporal} and its extension Mean Teacher~\cite{tarvainen2017mean} use an exponential moving average of previous predictions and model parameters to generate supervision targets for unlabeled data, respectively. 
Recently, the combination of consistency regularization and pseudo-labeling has gained more and more attention.
In MixMatch~\cite{berthelot2019mixmatch}, pseudo-labels are generated by averaging the predictions of different augmentations of the same sample.
ReMixMatch~\cite{berthelot2019remixmatch} and UDA~\cite{xie2020unsupervised} further extend this idea by dividing the set of augmentations into strong and weak augmentations, 
% Specifically, UDA~\cite{xie2020unsupervised} reveals that strong data augmentation is essential to improve SSL performance. 
while FixMatch~\cite{sohn2020fixmatch} leverages the model’s predictions on weakly-augmented unlabeled samples to create pseudo-labels for the strongly-augmented versions of the same samples.

% Of the aforementioned methods, CLS bears the closest resemblance to FixMatch~\cite{sohn2020fixmatch}, which is a state-of-the-art method that combines pseudo-labeling and consistency regularization.
% , where Pseudo-labeling encourages high-confidence predictions and consistency constraints is imposed between weakly-augmented samples and strongly-augmented samples.
However, all the methods mentioned above are self-labeling approaches.
An obvious obstacle in such methods is the confirmation bias~\cite{arazo2020pseudo}: the performance of self-labeling methods is restricted by the inaccurate artificial labels.
To  resolve this issue, our method generates adaptive sample weights for artificial labels, where samples with low confidence  are given small weights so that the effect of inaccurate labels can be effectively reduced. 
In addition, we borrow the idea of co-training~\cite{han2018co,yu2019does,ke2019dual} to exchange high-confidence labels between two collaborative networks.
% generate high-quality artificial labels for each other by learning two collaborative networks simultaneously.
Similar to UPS~\cite{rizve2021in}, CLS also generates complementary labels to improve robustness to inaccurate artificial labels.

% \subsection{Co-training}
%  Co-teaching~\cite{han2018co} proposed to train two collaborative networks simultaneously to select online clean data for each other to filter noisy labels, and Co-teaching+~\cite{yu2019does} further extended this method by updating the network on disagreement data to avoid two networks collapsing into each other.

% \subsection{summary}
% Contrary to all aforementioned methods, our method discards the labels that are highly likely to be
% noisy, and utilize the noisy samples as unlabeled data to regularize training in a SSL manner. Ding
% et al. (2018) and Kong et al. (2019) have shown that SSL method is effective in LNL. However, their
% methods do not perform well under high levels of noise, whereas our method can better distinguish
% and utilize noisy samples. Besides leveraging SSL, our method also introduces other advantages.
% Compared to self-training methods (Jiang et al., 2018; Arazo et al., 2019), our method can avoid
% the confirmation bias problem (Tarvainen  Valpola, 2017) by training two networks to filter error
% for each other. Compared to Co-teaching (Han et al., 2018) and Co-teaching+ (Yu et al., 2019), our
% method is more robust to noise by enabling the two networks to teach each other implicitly at each
% epoch (co-divide) and explicitly at each mini-batch (label co-refinement and co-guessing).

%---------------------------------------------------------------------------------------------------------
\section{Conclusion}
% In this paper, we proposed the Cross Labeling Supervision (CLS) method for SSL. 
% Key to CLS is the idea that one network learning from the supervision of another independent network can help alleviate the problem of confirmation bias.
% The learning process of CLS consists of two main steps: reweighting the artificial labels (both pseudo labels and complementary labels) according to their prediction confidence, and exchanging reliable artificial labels between the two different initialized networks for co-training.
% Experiments on standard SSL benchmarks show that CLS outperforms other state-of-the-art methods.
In this paper, we proposed the Cross Labeling Supervision (CLS) method for SSL. Key to CLS is the idea that one network learning from the supervision of another independent network can help alleviate the problem of confirmation bias. The learning process of CLS consists of two main steps: reweighting the artificial labels (both pseudo labels and complementary labels) according to their prediction confidence and exchanging reliable artificial labels between the two different initialized networks for co-training. Experiments on standard SSL benchmarks show that CLS outperforms other state-of-the-art methods.

\newpage
%%%%%%%%% REFERENCES
% {\small
% \bibliographystyle{ieee_fullname}
% \bibliography{main}
% }
\bibliographystyle{ACM-Reference-Format}
\bibliography{main}

\end{document}